\pdfoutput=1

\documentclass[11pt]{article}

\usepackage[final]{acl}

\usepackage{times}
\usepackage{latexsym}

\usepackage[T1]{fontenc}

\usepackage[utf8]{inputenc}

\usepackage{microtype}

\usepackage{inconsolata}

\usepackage{amssymb}
\usepackage{amsmath}
\usepackage{caption}
\usepackage{subcaption}
\usepackage{adjustbox}
\usepackage{multirow}
\usepackage{todonotes}
\usepackage{url}
\usepackage{hyperref}
\usepackage{array}
\usepackage{longtable}
\usepackage{tabularx}
\usepackage{booktabs}
\usepackage{ltablex}
\usepackage{fancyvrb}
\usepackage{float}
\usepackage{enumitem}
\usepackage{rotating}

\usepackage{amsmath}
\newcommand{\floor}[1]{\lfloor #1 \rfloor}
\newcommand{\ceil}[1]{\lceil #1 \rceil}

\usepackage{titlesec}
\titlespacing*{\paragraph}
  {0pt}    
  {5pt}   
  {5pt}    

\title{Do Factual Recall Mechanisms Carry over from Text to Speech in Multimodal Language Models?}

\def\authorsep{\hspace{0.3em}}

\author{Luca Modica\textsuperscript{1,3,4*} \authorsep Filip Landin\textsuperscript{2,3,4*} \authorsep Mehrdad Farahani\textsuperscript{3,4*} \authorsep Livia Qian\textsuperscript{5} \\ 
\textbf{Gabriel Skantze\textsuperscript{5}} \authorsep 
\textbf{Richard Johansson\textsuperscript{3,4}} \medskip\\
\null\textsuperscript{1}Zenseact \quad \null\textsuperscript{2}Unbox AI \quad
\null\textsuperscript{3}Chalmers University of Technology \\ \null\textsuperscript{4}University of Gothenburg \quad
\null\textsuperscript{5}KTH Royal Institute of Technology \medskip\\
\texttt{mehrdad.farahani@chalmers.se}}

\begin{document}
\maketitle

\def\thefootnote{*}\footnotetext{Equal contribution.}\def\thefootnote{\arabic{footnote}}
\renewcommand{\thefootnote}{\fnsymbol{footnote}}

\renewcommand{\thefootnote}{\arabic{footnote}}
\setcounter{footnote}{0}

\begin{abstract}
    In recent years, several Speech Language Models (SLMs) that represent speech and written text jointly have been presented. The question then emerges about how model-internal mechanisms are \emph{similar} and \emph{different} when operating in the two modalities. We focus on how these systems encode, store, and retrieve factual knowledge, which has previously been investigated for text-only models. 
    To investigate mechanisms behind the storage and recall of factual association in SLMs, we leverage Causal Mediation Analysis, a technique previously applied to text-based models.
    
    Initial results using \texttt{SpiritLM}, a multimodal model integrating discrete speech tokens reveal discrepancies between text-to-text and speech-to-text results, suggesting that the emergent mechanisms for factual recall are only partially carried over from the text to the speech modality. 
    %
    %
    %
    These results advance our understanding of how internal mechanisms encode factual associations in SLMs while contributing insights for improving speech-enabled AI systems.
\end{abstract}

\section{Introduction}

Large Language Models (LLMs) have demonstrated exceptional capabilities in various NLP tasks, including answering factual questions such as \textit{``the capital of Italy is''} by relying on information stored in their parameters \cite{petroni2019languagemodelsknowledgebases}. However, these systems still suffer from hallucination and are prone to committing factual errors, which limits their trustworthiness and usability \cite{kandpal2023largelanguagemodelsstruggle}: this motivates further investigation into the mechanisms behind knowledge recall and factual memory. Research using intervention-based methods reveals that factual knowledge can be \emph{localized} within text-based LLMs, particularly in mid-layer feed-forward networks (MLP) \cite{geva2021transformerfeedforwardlayerskeyvalue, meng2023locatingeditingfactualassociations,geva2023dissecting}. These findings are being used to develop methods that edit model parameters, allowing for precise intervention on factual associations, representing a step forward in more accurate and steerable models \cite{meng2023locatingeditingfactualassociations, meng2023memit}.  

Speech-language models trained directly on audio without text supervision -- such as those using GSLM-style training \cite{lakhotia2021generativespokenlanguagemodeling} -- have shown promise in speech understanding tasks \cite{lin2025surveymechanisticinterpretabilitymultimodal, basu2024understandinginformationstoragetransfer,peng2025surveyspeechlargelanguage,hassid2024textuallypretrainedspeechlanguage, zhang2023speechgptempoweringlargelanguage}. Since they do not leverage text-based knowledge, their factual understanding is more limited. On the other hand, speech models built on top of LLMs, like \texttt{SpiritLM} \cite{nguyen2024spiritlminterleavedspoken}, might retain or develop a deeper understanding of knowledge encoded in the text-based model. What is less understood is whether this behavior, if it exists, originates from the separate training on speech data or whether it comes from mechanisms learned from text. This opens up interesting research questions:
\begin{itemize}\addtolength{\itemsep}{-0.5\baselineskip}
    \item Are the mechanisms behind factual recall modality-independent?  
    \item Does factual localization in speech-based inputs emerge independently without reliance on the backbone architecture?
\end{itemize}

In this paper, we investigate where and how factual associations are stored and recalled in \texttt{SpiritLM} by using Causal Tracing (CT), one of the intervention-based techniques used to study the causal effect of components within a neural network \cite{meng2023locatingeditingfactualassociations}. We focus on two specific input-wise settings:
\begin{enumerate}\addtolength{\itemsep}{-0.5\baselineskip}
    \item T$\rightarrow$T (text-to-text): where the model receives and produces text. 
    \item S$\rightarrow$T (speech-to-text): where the model takes audio as input but still generates a text output.
\end{enumerate}
By extending CT to analyze factual recall in a multimodal setting, we show that speech input leads to weaker but detectable traces of factual localization.

\section{Methodology}
This section first describes the causal mediation analysis (CMA) framework and the mathematical underpinnings of causal tracing, followed by the \texttt{SpiritLM} model, dataset preparation, and the experiment design.

\subsection{Preliminaries: Causal Mediation Analysis} \label{subsection:prel-cma}
CMA is a framework for investigating questions about the relative contributions to an overall effect of individual components in a complex system \cite{pearl2013directindirecteffects}. Following \newcite{vig2020causalmediationanalysisinterpreting}, it has emerged as part of the standard toolbox for the analysis of LLMs; in mechanistic interpretability, it is also known as \emph{activation patching} \cite{heimersheim2024activation}.
\citet{meng2023locatingeditingfactualassociations} applied CMA to investigate factual recall in LMs. Their approach consists of three steps:

\paragraph{Clean run.} The LM is provided a clean prompt $X = x$, producing a probability $\mathbb{P}_{x}[o]$, where $o$ denotes the expected decoded  token. The corresponding hidden states from this inference are cached.

\paragraph{Corrupted run.} The model receives a corrupted input prompt $X = x^*$, resulting in a new predicted output probability $\mathbb{P}_{x^*}[o]$. \newcite{meng2023locatingeditingfactualassociations} carried out the corruption intervention by obfuscating the subject tokens with noise proportional to the standard deviation over all input embeddings.

\paragraph{Corrupted-with-restoration run.} The same corrupted prompt $X = x^*$ is passed to the model, but with the activation value of selected component $C_i$ restored (patched) from the \textit{clean run}. The result is denoted $\mathbb{P}_{x^*, \text{clean } C_i}[o]$, where "$\text{clean } C_i$" refers to the value of the component $C_i$ from the clean inference.

\vspace{1.5mm}

The results of the three runs allow us to quantify 
the mediated effects of interventions. 
%
The relative contribution of a hidden-state mediator is measured by the \textit{Indirect Effect} (IE), defined as the difference between the corrupted-with-restoration run and the corrupted run:
\begin{equation*}
    \text{IE} = \mathbb{P}_{x^*, \text{clean } C_i}[o] - \mathbb{P}_{x^*}[o].
\end{equation*}
By averaging over multiple prompts, we obtain Average Indirect Effect (AIE) at different levels of the model components and then visualize the contribution results.

\subsection{The multimodal large language model under study: The \texttt{SpiritLM} model}
\label{spiritlm-section}
Our work examines \texttt{SpiritLM} \cite{nguyen2024spiritlminterleavedspoken} as a case of a multimodal (speech) language model that can generate text and audio language content. Furthermore, \texttt{SpiritLM} uses discrete speech tokens and is trained on interleaved speech and text token sequences for better generalization and alignment across modalities, making it well-suited for the proposed study. 

\begin{figure}
    \centering
    \includegraphics[width=\linewidth]{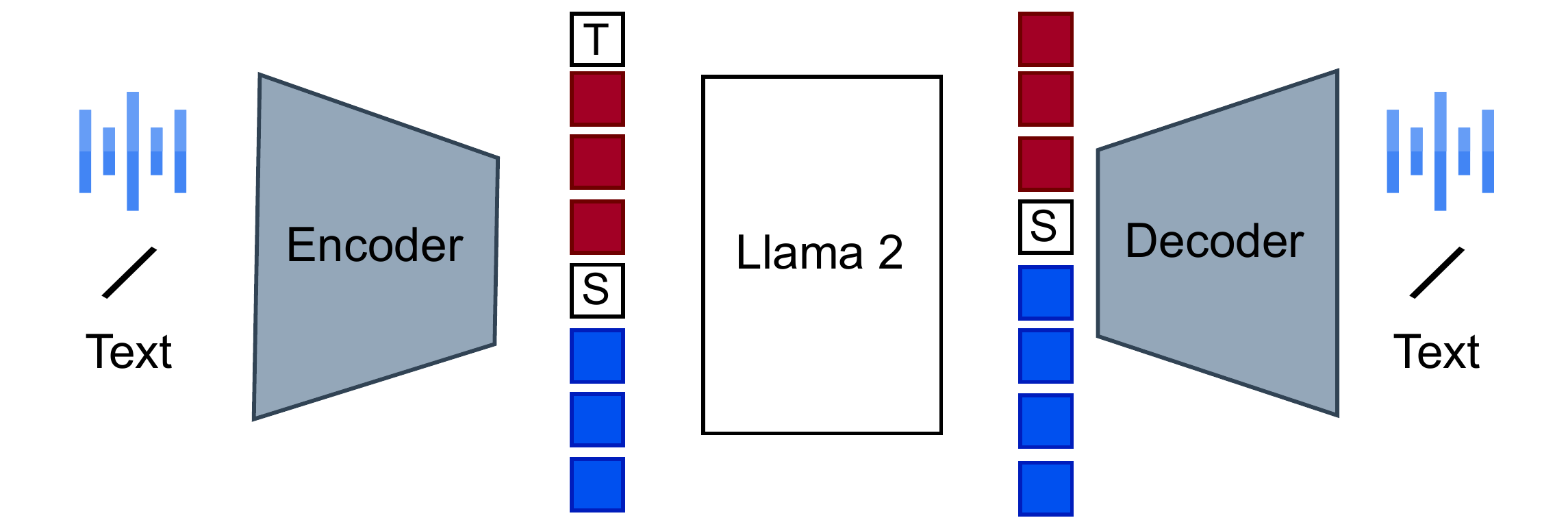}
    \caption{The SpiritLM architecture.}
    \label{fig:spiritlm}
\end{figure}

We illustrate the high-level architecture of \texttt{SpiritLM} in Figure \ref{fig:spiritlm}. The model handles mixed speech and text inputs using special \textit{modality declaration tokens} ("T" for text, "S" for speech). Audio is discretized into tokens with HuBERT \cite{hsu2021hubertselfsupervisedspeechrepresentation} and text with the Llama2 tokenizer. The interleaved sequence, with each portion prefixed by its respective modality tokens, is input to Llama2. The model predicts the next tokens based on the most recent modality token: a "T" token prompts text generation, while "S" prompts discrete speech tokens. At inference, speech tokens are decoded with HiFi-GAN~\cite{kong2020hifigangenerativeadversarialnetworks}.

The employed speech representation allows targeting and analyzing specific speech tokens in a Causal Mediation Analysis experiment in both a uni-modal and cross-modal context.

\subsection{Dataset and data preparation} \label{data-prep}
For our study, we use the \textit{Known} dataset \cite{meng2023locatingeditingfactualassociations}: It contains almost 1000 factual prompts that the GPT2-XL model knows,\footnote{Link to the collection of factual prompts: \url{https://rome.baulab.info/data/dsets/known_1000.json}}
with the annotated subject and object (expected correct answer).

Starting from the available text data, we introduce the speech modality counterpart for each information in Known (prompt, subject, and object). The utterances are generated using the TTS model MeloTTS \cite{zhao2024melo}, which is based on architectures that leverage adversarial learning to improve expressive power and high-quality speech synthesis \cite{kim2021conditionalvariationalautoencoderadversarial, kong2023vits2improvingqualityefficiency}. We assess the reliability of the curated speech modality through two complementary approaches: manual inspection of challenging samples, particularly prompts, and automatic transcription of the generated audio using \textit{Whisper-small},\footnote{Link to the model checkpoint used: \url{https://huggingface.co/openai/whisper-small}} a lightweight ASR model. The prompt transcription results yield a Word Error Rate of $19\%$: this demonstrates good performance despite the inherent difficulty of transcribing proper nouns, and the reliability of the TTS model.

To further ensure the dataset quality for the subsequent experiment, we filter the original collection factual statements based on the model performance in the 2 different modalities as input, resulting in 2 datasets: \textit{Known-t2t} and \textit{Known-s2t}. \textit{Known-t2t} includes datapoints where the model readily generates either an exact correct answer or a close variant in a $\text{Text} \rightarrow \text{Text}$ scenario. For instance, "Rome" is correct for the prompt "The capital of Italy is \_\_\_", while answers like "Rome, Italy" or "the city of Rome" are considered partially correct. \textit{Known-s2t} follows the same selection criteria, but in a $\text{Speech} \rightarrow \text{Text}$ setting. 


\subsection{Experiment design}
Factual associations in \texttt{SpiritLM} are investigated through two CMA experiments, in text and speech domain, in order to determine causal effects of network components: single transformer layers, MLP, and attention sub-layers.

Experiments are conducted on prompts from the datasets introduced in \ref{data-prep}. Similarly to \citet{meng2023locatingeditingfactualassociations}, the corrupted run is done by adding noise to the representation of the subject tokens.

\paragraph{Experiment 1: Within-modality factual recall (Text $\rightarrow$ Text)}
In the first experiment, a text prompt is fed into the model and the log probability of predicting the corresponding attribute is computed for each of the three CMA iterations -- clean, corrupted, and corrupted-with-restoration runs -- described in Section \ref{subsection:prel-cma}. The IE is aggregated by the position of the token in the sentence: \textit{first subject token, middle subject tokens, last subject token, first subsequent token, further tokens, and last token}, averaged over all prompts (AIE), and presented as log AIE for readability and comparison.

\paragraph{Experiment 2: Cross-modality factual recall (Speech $\rightarrow$  Text)} 
The second experiment is similar, but uses the synthesized version of the dataset, where prompts are converted to audio. Each utterance is encoded and discretized by \texttt{HuBERT}), and the resulting tokens are fed into the language model, where the CMA pipeline is applied as in the previous experiment. 
An additional challenge here is that, in the corrupted run, it is no longer obvious how to localize the  subject token(s) in the input prompt. Connectionist Temporal Classification (CTC)-based forced alignment \cite{K_rzinger_2020} is therefore used to find the target time range of the subject in the utterance, and thus the related range of the speech tokens (see Appendix \ref{app:alignment} for more details). The same technique admits a mapping between speech tokens and the corresponding text ones, which is used to post-process the CMA results. The causal traces of the speech (HuBERT) token are aggregated, as for text, by the corresponding text tokens, which facilitates direct comparison and interpretation of causal influence across modalities. The quality of the forced alignment is validated by manually inspecting the speech segments corresponding to text tokens, ensuring that token boundaries are properly aligned without overlaps or significant gaps throughout the prompt utterance.

\begin{figure*}[!ht]
\centering
\setlength{\tabcolsep}{2pt}
\renewcommand{\arraystretch}{1.2}

\begin{tabular}{c c c c}
& \multicolumn{1}{c}{\small \textbf{\shortstack{single \\ patched layer}}} &
  \multicolumn{1}{c}{\small \textbf{\shortstack{center of interval \\ of 5 patched mlp layers}}} &
  \multicolumn{1}{c}{\small \textbf{\shortstack{center of interval \\ of 5 patched attention layers}}} \\[0.3em]

\multirow{1}{*}[0.1em]{\rotatebox[origin=c]{90}{\shortstack{\scriptsize \textbf{T $\rightarrow$ T} \\ \scriptsize (LLaMA 2.7B)}}} &
\adjustbox{valign=m}{%
  \begin{minipage}{0.31\textwidth}
    \small (a)\\
    \includegraphics[width=0.95\linewidth]{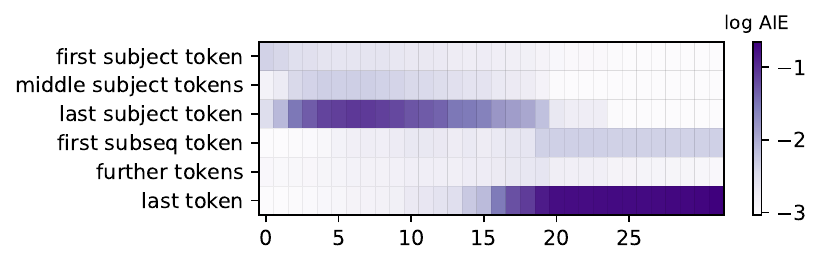}
  \end{minipage}
} &
\adjustbox{valign=m}{%
  \begin{minipage}{0.31\textwidth}
    \small (b)\\
    \includegraphics[width=0.95\linewidth]{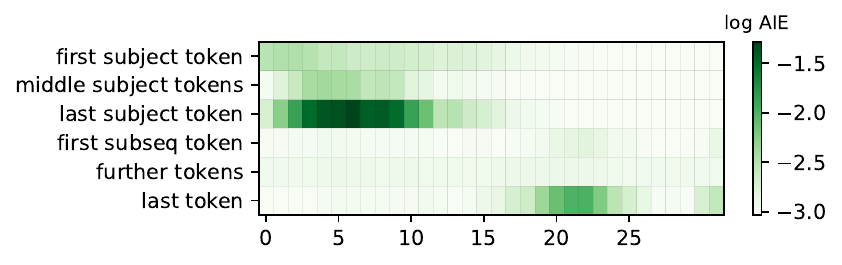}
  \end{minipage}
} &
\adjustbox{valign=m}{%
  \begin{minipage}{0.31\textwidth}
    \small (c)\\
    \includegraphics[width=0.95\linewidth]{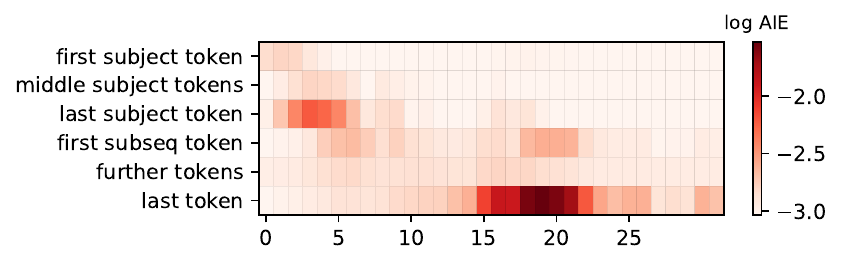}
  \end{minipage}
} \\

\multirow{1}{*}[0.1em]{\rotatebox[origin=c]{90}{\shortstack{\scriptsize \textbf{T $\rightarrow$ T} \\ \scriptsize (SpiritLM)}}} &
\adjustbox{valign=m}{%
  \begin{minipage}{0.31\textwidth}
    \small (d)\\
    \includegraphics[width=0.95\linewidth]{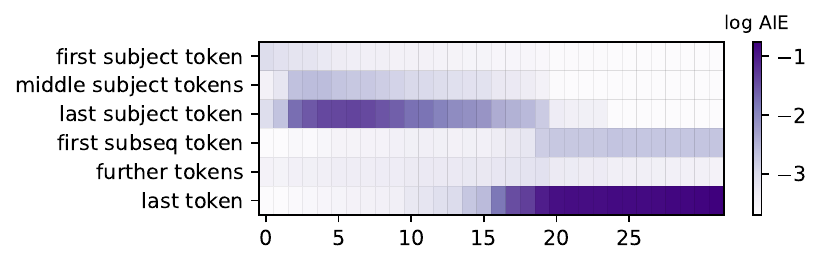}
  \end{minipage}
} &
\adjustbox{valign=m}{%
  \begin{minipage}{0.31\textwidth}
    \small (e)\\
    \includegraphics[width=0.95\linewidth]{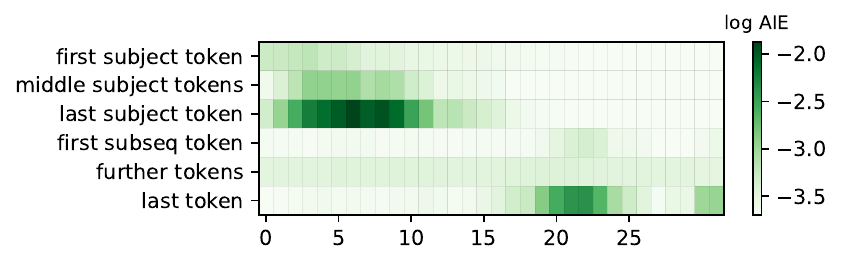}
  \end{minipage}
} &
\adjustbox{valign=m}{%
  \begin{minipage}{0.31\textwidth}
    \small (f)\\
    \includegraphics[width=0.95\linewidth]{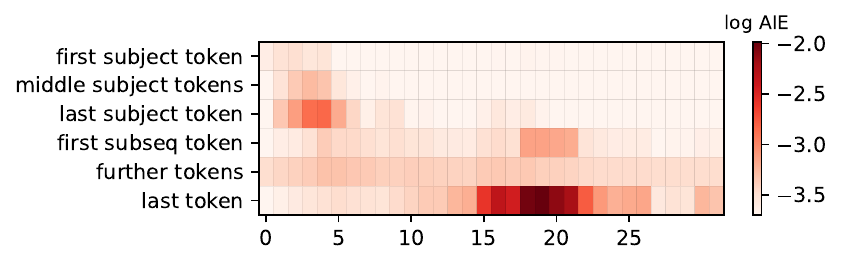}
  \end{minipage}
} \\

\multirow{1}{*}[0.1em]{\rotatebox[origin=c]{90}{\shortstack{\scriptsize \textbf{S $\rightarrow$ T} \\ \scriptsize (SpiritLM)}}} &
\adjustbox{valign=m}{%
  \begin{minipage}{0.31\textwidth}
    \small (g)\\
    \includegraphics[width=0.95\linewidth]{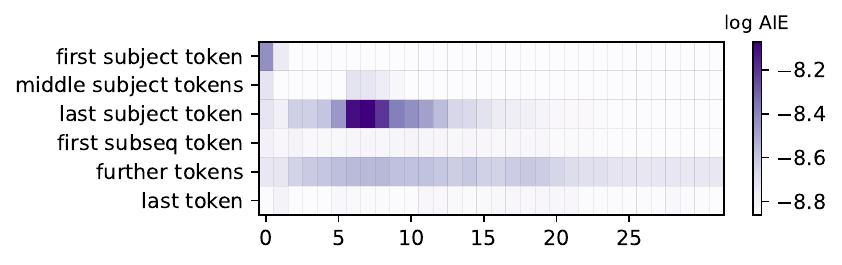}
  \end{minipage}
} &
\adjustbox{valign=m}{%
  \begin{minipage}{0.31\textwidth}
    \small (h)\\
    \includegraphics[width=0.95\linewidth]{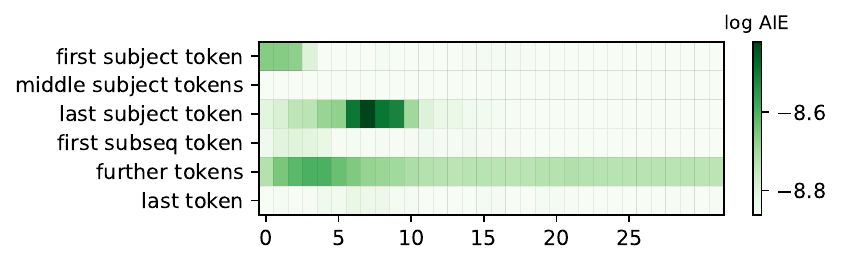}
  \end{minipage}
} &
\adjustbox{valign=m}{%
  \begin{minipage}{0.31\textwidth}
    \small (i)\\
    \includegraphics[width=0.95\linewidth]{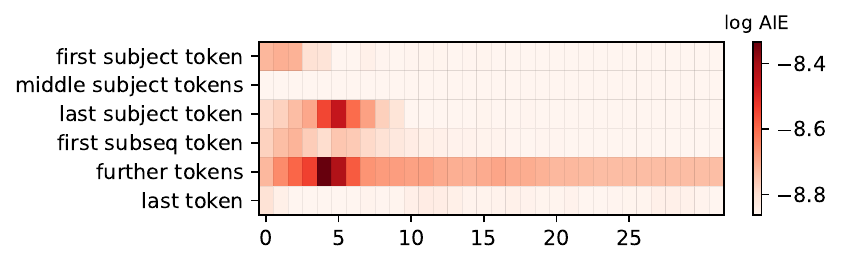}
  \end{minipage}
} \\
\end{tabular}

\vspace{1em}
\caption{Log-scaled AIE across different modules and modalities over 754 prompts. In each subfigure, the x-axis represents the layers and the y-axis shows the tokens of interest.}
\label{fig:avg_indirect_effect_3e_log}
\end{figure*}

\section{Results and Discussion} 
We begin our experiments at two levels to examine whether factual associations -- previously shown to localize around subject tokens in text-only models -- can also be recalled and expressed in other modalities. In our case, we focus on \texttt{SpiritLM} and its speech modality. To measure this, we use causal mediation analysis to compute the Average Indirect Effect (AIE) for each layer and token across all filtered query prompts. The AIE captures the marginal contribution of an internal component to the final factual prediction under intervention. Higher AIE values indicate which layers and positions influence factual recall more.

As a baseline, we perform CT on the backbone model used in \texttt{SpiritLM}, using text-to-text prompts. As expected from prior work by \newcite{meng2023locatingeditingfactualassociations}, we observe strong causal signals (AIE) centered on subject tokens at early layers, especially in mid-layer MLPs (see Figure~\ref{fig:avg_indirect_effect_3e_log}). We also detect notable effects at the final token position in upper layers, where strong causality is typically observed. Extending the same CT analysis to \texttt{SpiritLM} in the T$\rightarrow$T setting reveals nearly identical behavior: consistent causal signals around subject tokens across hidden States, MLP, and attention layers (Figure~\ref{fig:avg_indirect_effect_3e_log}). This confirms that text-processing pathways in \texttt{SpiritLM} preserve their original capabilities after fine-tuning on speech.

On the other hand, we observe different results when the input is speech (S$\rightarrow$T). The AIE drops drastically, showing a much more diffuse and lower-magnitude signal; however, we can still observe an effect around the subject tokens in the MLP and attention layers (see Figure~\ref{fig:avg_indirect_effect_3e_log}).

Our experiments suggest that factual associations in spoken language models like \texttt{SpiritLM} are not strictly modality-dependent. Although the model is capable of retrieving knowledge from both modalities -- including, to some extent, speech -- the 
factual recall mechanisms are much more readily activated when the model is provided with text input than with speech.
For \texttt{SpiritLM}, text serves as a more structured and reliable trigger for recalling facts, suggesting that the speech-based fine-tuning in this model does not fully utilize the fact-recalling mechanisms learned by the text-based backbone model.
However, we do not have sufficient evidence to conclude whether the partial transfer of factual capabilities from text to speech in \texttt{SpiritLM} arises from noise introduced by controlled conditions or limitations of the mapping between speech and text tokens. Based on recent studies \cite{xiang2025understandingmodalitygapempirical, cuervo2026closinggaptextspeech}, we also hypothesize that a likely cause is the semantic gap between the two modalities, possibly arising from the post-training speech adaptation of \texttt{SpiritLM}'s text-only backbone: while the two modalities may become increasingly aligned in direction (cosine similarity) across deeper layers, a divergence in magnitude (Euclidean distance) can still persist, ultimately compromising the transfer of factual knowledge to the speech modality.

\section{Related Work}


Interpretability for speech LLMs remains underexplored compared to their text counterparts: in particular, investigating to what extent multimodal text/speech systems share underlying mechanisms remains unexplored. 
Recent works \cite{pasad2024selfsupervisedspeechmodelsknow,pasad2022layerwiseanalysisselfsupervisedspeech,shen2024encoding} have explored speech model interpretability, considering 
speech features at different granularity levels (e.g., word boundaries, pronunciation), finding, for example, that frame-level representations within each word segment are not all equally informative. These studies lead to open questions more related to how sentence-level properties (e.g., subject) are encoded, a gap our work seeks to fill. More recently, \citet{glazer2025transcriptionmechanisticinterpretabilityasr} explored mechanistic interpretability for ASR systems, applying logit lens and activation patching to reveal internal model dynamics responsible for repetition hallucinations and semantic biases within acoustic representations: these findings suggest promising directions for similar investigations in speech LLMs and beyond the ASR setting.

\section{Conclusion}
The study investigates the mechanisms of factual recall in Speech LLMs, focusing on whether this process in the speech modality operates independently or relies on the text modality and the capabilities of the original text model. By using the CMA framework with \texttt{SpiritLM}, we show that the model preserves the same text-based computation pathways as its corresponding text-only counterpart, while the speech modality leads to a considerably weaker causal effect at the level of the MLP and attention layers. Although the latter does not conclusively prove if knowledge from text is transferred to the speech modality, these preliminary insights hint that speech LLMs are not strictly modality-dependent when recalling facts.

\section*{Limitations and Future Directions}
\textbf{Dataset selectivity.} We conducted our experiment on a single synthesized speech dataset -- \textit{Known} \cite{meng2023locatingeditingfactualassociations}, which might not capture all the nuances of how modality interactions affect factual memory tracing. Using other datasets, from \textit{Spoken SQuAD} \cite{li2018spokensquadstudymitigating} to a synthesized version of \textit{PopQA} \cite{mallen2023trustlanguagemodelsinvestigating}, can provide a valuable contribution in this field.

\textbf{Model generalization.} Although the insights obtained using \texttt{SpiritLM}, their compatibility with other Speech LLMs that employ discrete speech tokens needs to be explored \cite{zhang2023speechgptempoweringlargelanguage, rubenstein2023audiopalmlargelanguagemodel}. Furthermore, replicating our experiments within \texttt{SpiritLM} through different text-only backbones \cite{yang2025qwen3technicalreport, jiang2023mistral7b} or a joint speech-text training strategy would be crucial to assess the generalizability of our findings.

\textbf{Discrete speech tokens limitations.} Using discrete tokens represents an interesting strategy to integrate the modality with text-based tokens seamlessly; however, recent studies have shown performance limitations on semantic understanding tasks, which might affect results of factual recall studies of speech Large Language Models \cite{wang-etal-2025-speech}. Considering speech LLMs that employ different strategies to convey a richer speech representation, such as training on continuous speech representation \cite{peng2025surveyspeechlargelanguage, tang2024salmonngenerichearingabilities}, can represent a promising future direction of this investigation.

\section*{Ethical Considerations}
This study focuses on the interpretability of speech-language models. As part of our research, we do not release any new models or datasets; therefore, we do not implicate any potential risks or concerns related to the misuse of our results.

\section*{Acknowledgments}
This research was funded by the Wallenberg AI, Autonomous Systems and Software
Program (WASP) funded by the Knut and Alice Wallenberg Foundation. The computations were
enabled by resources provided by the National Academic Infrastructure for Supercomputing in
Sweden (NAISS) at Alvis partly funded by the Swedish Research Council through grant agreement no. 2022-06725.
We also acknowledge the Computer Science and Engineering department at Chalmers and the University of Gothenburg, which funds part of the conference costs through the \emph{Lars Pareto travel grant}.

\bibliography{custom}

\clearpage
\newpage
\appendix
\onecolumn
\section{Forced Alignment for Cross-modal Token Mapping: Implementation Details}
\label{app:alignment}

\paragraph{Text preprocessing for CTC.} For the transcription to be compatible with the forced alignment, a text preprocessing is necessary to ensure all characters are included in the CTC model vocabulary. For example, digits and special characters such as "\%" are converted to their written format (e.g, "0" becomes "zero", or "\%" becomes "percent"). On the text token-level, preprocessing can lead to a longer or shorter sequence of tokens, compared to the original text being tokenized. We later refer to the preprocessed text tokens as \textit{spoken text tokens}. The preprocessing step concludes by joining the spoken text tokens into a single string, using the word boundary character defined by the CTC model as a separator.

\paragraph{Frame-wise label probability estimation from audio waveform.} We generate emission probabilities per audio frame, using the pre-trained \texttt{HuBERT-LARGE}\footnote{Link to the model checkpoint used: \url{https://huggingface.co/facebook/hubert-large-ls960-ft}} model as a speech tokenizer. This model is fine-tuned for automatic speech recognition (ASR) with CTC loss, representing a suitable candidate for this use case \cite{hsu2021hubertselfsupervisedspeechrepresentation}.

\paragraph{Trellis matrix generation with log-probability of label alignments at each time step.} Given an audio input sequence $\mathbf{X} = (x_1, \ldots, x_T)$ and transcript labels $(c_1, \ldots, c_N)$ at the character level, we compute through dynamic programming and map all possible joint probabilities in the trellis diagram matrix $K \in \mathbb{R}^{T \times N}$; $K_{(t, j)}$ represents the maximum log-probability of aligning the first labels $j$ up to time $t$. To compute the probability at time step $t+1$ for label $c_{j+1}$, we consider two possible transitions: either we stayed on the same label $c_{j+1}$ or transitioned from $c_j$ to $c_{j+1}$. Based on these criteria, the trellis is updated as follows:
\begin{equation*}
    K_{(t+1, j+1)} = \max 
    \begin{cases}
        K_{(t, j)} \, p(t+1, c_{j+1}) \\
        K_{(t, j+1)} \, p(t+1, \text{repeat})
    \end{cases}
\end{equation*}

where $p(t+1, c_{j+1})$ is the probability of emitting label $c_{j+1}$ at time $t+1$, and $p(t+1, \text{repeat})$ is the probability of emitting no label change.
    
\paragraph{Find the most likely path from the trellis matrix.} Once the trellis is generated, we will traverse it following the elements with the highest probability. Starting from the last label index belonging to the last time step, we progress in the matrix backwards, choosing to keep the current label or move to the previous label based on the highest probability for each time step. The process ends when we reach the beginning of the sequence, obtaining the most likely path that aligns text and audio.

\paragraph{Merge repetitions and segments into words (spoken text tokens).}
The final step involves postprocessing the output from the optimal path. Because the path may contain consecutive repetitions of the same label, we merge path points corresponding to repeated characters into a single segment to make it close to the original transcript.\footnote{When merging path points into a single segment, we use the average probability of all frames in that segment.}
Similarly, we group segments that correspond to the same spoken text token, using the word boundary character as a guide. The result is a sequence of segments, each representing a spoken text token from the transcript and annotated with the corresponding range of audio frames and average emission probability. \\

\begin{figure}
    \centering
    \includegraphics[width=1\linewidth]{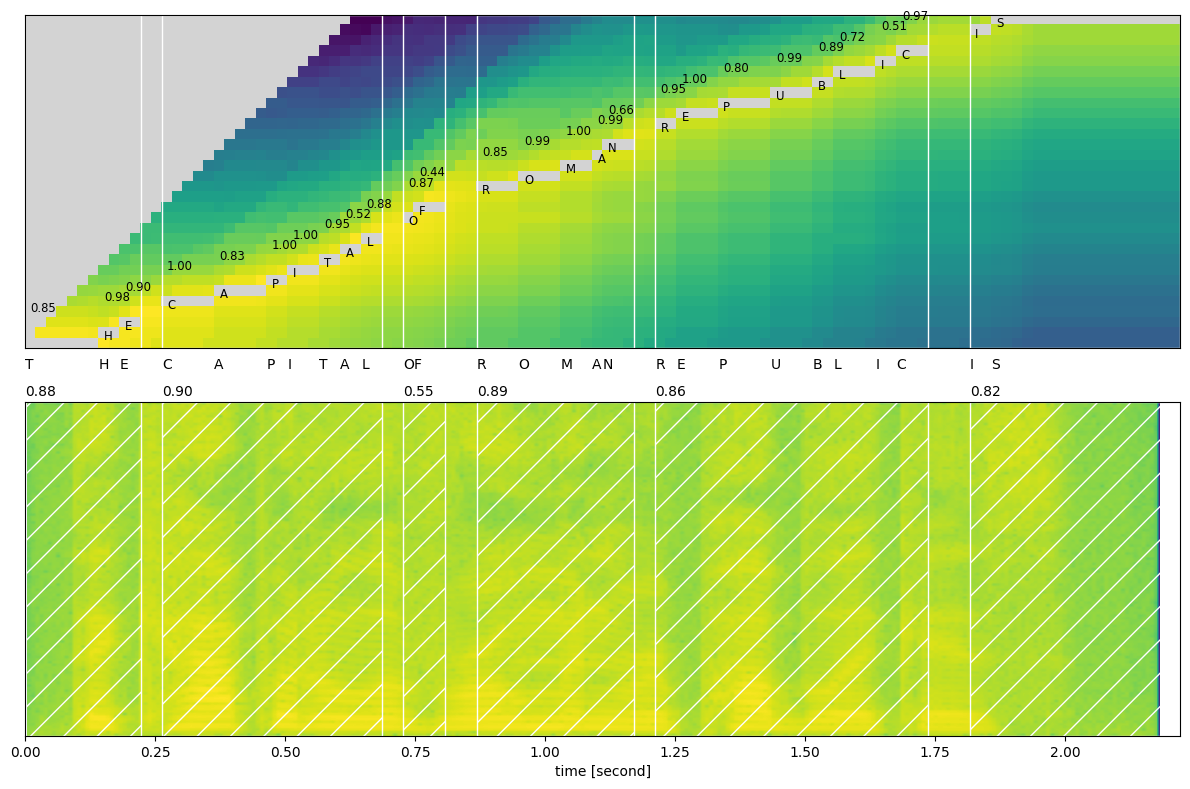}
    \caption{Results of the forced alignment for a speech utterance (transcript: "The capital of Roman Republic is"). The plot on top shows the trellis matrix, with the highlighted optimal path and score for each labeled letter; on the bottom, instead, we show the mel spectrogram of the spoken utterance, with the corresponding boundaries between each (spoken) text token.}
    \label{fig:result-alignment}
\end{figure}

From this segmentation, we can derive the time range of each text token, which can also be used to obtain the speech token range, using the token rate of the speech tokenizer. The last result allows us to directly map text tokens and related speech tokens. Figure \ref{fig:result-alignment} illustrates an example of the final output from CTC-based forced alignment, demonstrating this alignment process. Beginning with the preprocessed transcript "THE|CAPITAL|OF|ROMAN|REPUBLIC|IS," we align each text token with its corresponding segment in the audio. For each text token, we label the aligned speech segment with the average probability over the merged segment, clearly indicating its position within the utterance as a highlighted segment on the spectrogram, with boundaries marking its start and end. This segmentation process allows us to determine the precise time range for each text token.
\begin{itemize}
    \item Given the frame range $(f_{start}, f_{end})$ of a spoken text token segment, we can first compute its time range in seconds $(s_{start}, s_{end})$ in the utterance with the following formula:
    \begin{equation}
        s_{start} = \frac{\floor{ratio \cdot f_{start}}}{sr}, 
        s_{end} = \frac{\floor{ratio \cdot f_{end}}}{sr},
    \end{equation}
    Where $sr$ represents the sample rate of the original sampled waveform $Z = (z_1, \ldots, z_M)$, while $ratio = \frac{M}{T}$ represents the number of samples contained in a frame.

    \item Then, considering the token rate of the speech tokenizer $tr$ and the previously computed time range $(s_{start}, s_{end})$, the corresponding speech token range $(stk_{start}, stk_{end})$ is given by:
    \begin{equation}
        stk_{start} = \floor{s_{start} \cdot tr}, stk_{end} = \ceil{s_{end} \cdot tr}.
    \end{equation}
\end{itemize}

This direct mapping provides a clear correspondence between each text token and its associated speech tokens, linking elements of the transcript to their acoustic realizations in the audio.

\end{document}